# Intent-First Aerial V2V for Tactical Coordination and Separation: Protocol and Performance Under Density and Disturbance

Mehrnaz Sabet, *Member, IEEE*

***Abstract*— Dense low-altitude aerial operations will require more than pre-flight route coordination and last-resort collision avoidance. Once aircraft are airborne, disturbances can emerge on timescales shorter than strategic reauthorization can reliably absorb, while collision avoidance is too late and too disruptive to serve as routine traffic management. Although tactical separation is recognized as the intermediate layer, realizing it at scale requires a deployable neighborhood communication mechanism that provides fresh, trusted, and actionable information for local coordination. This paper presents what is, to our knowledge, the first controller-coupled characterization of an all-airborne, sidelink-class, intent-first vehicle-to-vehicle (V2V) tactical neighborhood exchange stack for dense Unmanned Aircraft System Traffic Management (UTM) operations. Unlike awareness-only broadcast, the proposed exchange combines continuously refreshed state and intent beacons for local awareness, cooperative perception, and degraded-mode assessment with event-triggered messages for explicit yielding, sequencing, release, and contingency coordination. We implement and evaluate this model on an all-airborne V2V stack using sidelink-class C-V2X modules with authenticated freshness checks. Evaluation uses a scenario-driven, high-volume stress campaign supported by a real-time, field-anchored infrastructure. Results show that V2V plays specific operational roles: reducing stale-belief divergence, preserving observability through cooperative perception, rejecting invalid tactical messages, suppressing false local inference, and structuring shared-resource coordination. The implemented stack provides a viable communication layer for tactical separation in lower-to-moderate density regimes, but transitions toward guarded fallback as density, impairment, and interaction complexity increase. These findings position intent-first aerial V2V as a bounded, safety-critical enabler for scaling tactical coordination in disturbance-driven urban airspace.**

This work was supported in part by the National Aeronautics and Space Administration (NASA) under Grant 80NSSC24K0615 and in part by the National Artificial Intelligence Research Resource (NAIRR) Pilot under Award NAIRR240493. Any opinions, findings, conclusions, or recommendations expressed in this material are those of the author and do not necessarily reflect the views of NASA.

Mehrnaz Sabet is with Cornell University, Ithaca, NY 14853 USA (e-mail: ms3662@cornell.edu).

## I. INTRODUCTION

Future low-altitude airspace will become part of intelligent transportation systems. Aerial package delivery, infrastructure inspection, emergency response, public-safety operations, and emerging electric air-taxi services all point toward a future in which many automated or highly automated Unmanned Aircraft Systems (UAS) operate near cities, terminals, pickup/dropoff zones, and other shared resources. Like dense connected road traffic, these operations require independently operated vehicles to coordinate while preserving safety and flow. Unlike road traffic, however, low-altitude aerial traffic is three-dimensional, highly dynamic, more exposed to communication and sensing degradation, and less constrained by fixed lanes or road geometry. Scaling such operations therefore requires more than individual vehicle autonomy: it requires cooperative airspace mechanisms that can maintain local coordination under uncertainty.

Current UAS Traffic Management (UTM) architectures appropriately emphasize strategic deconfliction: allowing UAS operators and service providers to share planned flight paths, reserve or coordinate 4D operational volumes to avoid planned conflicts before departure and monitor conformance during flight. This layer is essential for interoperability and macro-scale flow management, but dense urban operations are not governed only by pre-planned conflicts. Once aircraft are airborne, local disturbances such as pad delays, wave-offs, pop-up constraints, navigation-integrity events, communication impairment, degraded sensing, contingency maneuvers, or uncooperative aircraft can invalidate strategic assumptions. At the other end of the stack, onboard collision avoidance provides the last-resort response when separation is already at risk. It is necessary, but it is too late and too disruptive to serve as routine traffic management. This motivates the intermediate tactical separation layer: an in-flight function that coordinates nearby aircraft to repair short-horizon conflicts before they become collision-avoidance events or broad strategic reauthorization cascades.

Although tactical separation is recognized in UTM conflict-management concepts, the communication mechanism needed to realize it at scale remains under-specified. Tactical separation requires nearby aircraft to exchange fresh, trusted, intent-bearing information and explicit coordination state quickly enough for local controllers to act before encounters collapse into collision-avoidance events. Without such exchange, tactical separation remains an assumed capability rather than a deployable coordination layer. This paper addresses that gap by designing and evaluating an all-airborne, intent-first V2V tactical neighborhood exchange for scalable local coordination.

The exchange has two channels. A periodic state and intent beacon maintains continuously refreshed local awareness, short-horizon intent, freshness, observability quality, integrity state, and maneuver authority. This channel supports awareness continuity, cooperative perception, degraded-mode assessment, and predictive local conflict evaluation. An event-triggered coordination channel is used only when aircraft need to assert or negotiate a specific local interaction outcome, such as yielding, commitment, release, rejoin, hotspot sequencing, or contingency protected-space behavior. The central claim is that aerial V2V becomes valuable for tactical separation when it functions as a bounded, trusted, freshness-aware coordination layer coupled to degraded-mode control, not as an awareness-only communication channel.

This paper implements and evaluates this model on an all-airborne V2V stack using sidelink-class C-V2X/PC5 modules. The stack is evaluated through a scenario-driven, high-volume stress campaign supported by a real-time, field-anchored evaluation infrastructure combining physical flight, hardware-, software-, and communications-in-the-loop testing, hybrid real-time operation, and scaled field-anchored simulation. The evaluation compares strategic-only coordination, non-V2V tactical coordination, and deployable V2V-enabled tactical coordination. This baseline structure allows us to distinguish the value of tactical coordination from the specific role and operational cost of realizing tactical coordination through deployable V2V exchange.

The evaluation asks (I) what operational roles intent-first aerial V2V plays beyond strategic-only or non-V2V tactical coordination, (II) where those benefits persist across density, impairment, observability, integrity, and shared-resource stress, and (III) what operational cost appears when tactical coordination is constrained by deployable V2V freshness, trust, and congestion limits.

These questions are answered through four contributions that separate the implemented V2V stack, the protocol design, the field-anchored evaluation methodology, and the resulting mechanism-level operating envelope:

1. All-airborne V2V implementation: We implement what is, to our knowledge, the first all-airborne V2V tactical-neighborhood exchange stack coupled to a tactical coordination controller. The implementation uses sidelink-class C-V2X/PC5 modules, authenticated freshness checks, and no ground relay, enabling field-anchored evaluation of tactical neighborhood exchange.

2. Two-channel tactical V2V protocol: We define a neighborhood exchange model that separates periodic state and intent beacons for local belief, cooperative perception, and degraded-mode assessment from event-triggered messages for yielding, commitment, release, sequencing, rejoin, and contingency coordination.

3. Field-anchored communication-control evaluation: We evaluate the stack through a scenario-driven stress campaign that couples PRR, latency, deadline misses, and freshness to closed-loop tactical outcomes across impairment, delayed context, integrity faults, degraded observability, mixed equipage, and hotspot interactions, treating communication metrics as inputs to transportation performance rather than as standalone link metrics.

4. Mechanism-level operating-envelope findings: We show that the evaluated intent-first aerial V2V stack provides mechanism-specific value, supporting stale-belief reduction, trusted coordination, false-inference suppression, cooperative perception, shared-resource sequencing, and partial-participation coordination, while identifying regimes where the system transitions toward guarded or backstop-heavy behavior.

The remainder of the paper is organized as follows. Section II reviews related work in UTM, V2X-enabled transportation control, aerial V2V/A2X, and cooperative perception. Section III presents the intent-first V2V coordination protocol. Section IV describes the evaluation methodology. Section V characterizes the communication envelope under density and impairment. Section VI presents controller-coupled operational results. Section VII discusses implications and limitations. Section VIII concludes.

## II. Related Work and Positioning

### *A. UTM Conflict Management and Tactical Communication*

UTM concepts of operations establish a layered conflict-management model for low-altitude UAS operations: strategic deconfliction coordinates planned operations through intent sharing, 4D operational-volume coordination, interoperability, constraint management, and pre-flight conflict prevention, with in-flight conformance monitoring and reauthorization where needed; tactical separation addresses short-horizon in-flight conflicts and local disturbances; and onboard collision avoidance provides the last-resort response when separation is already at risk [1]–[4]. UTM architectures define the strategic layer at a level that supports deployment-oriented interoperability and field testing. While the tactical layer is recognized, the deployable communication mechanism needed to realize it at scale remains under-specified: current concepts do not define the air-to-air channel, message schema, freshness rules, authentication model, or local coordination semantics needed to realize tactical separation as a controller-coupled service.

Awareness and conspicuity links are often discussed as communication mechanisms for improving cooperative visibility in low-altitude airspace and supporting detect-and-avoid [5], but they do not close the tactical coordination communication gap. ADS-B [6] and Remote ID [7] provide surveillance, identification, or accountability functions, but they were not designed to express short-horizon intent, observability state, tactical freshness windows, authenticated commitments, or event-triggered coordination primitives. Prior work has explored extending Remote ID-like messages for collision avoidance, which reinforces that baseline awareness formats are insufficient for neighborhood-scale tactical coordination [7]–[9].

Algorithmic work on tactical or self-separation has developed decision-theoretic, game-theoretic, self-organizing, and decentralized trajectory-planning approaches under assumed or modeled information exchange [10]–[12]. MADER/RMADER explicitly exchange planned trajectories and consider delay or asynchrony [13], [14]. These works show why communication matters for multi-agent safety, but they generally treat the network as an exchange substrate rather than characterizing a deployable all-airborne V2V tactical service with measured link behavior and closed-loop transportation outcomes.

### *B. Aerial V2V, A2X, and C-V2X*

Prior aerial V2V and A2X work establishes feasibility and standards direction, but it does not yet provide the controller-coupled tactical-neighborhood characterization addressed here. Several studies use UAVs as aerial infrastructure for ground V2X, such as relays, aerial base stations, or roadside-unit substitutes [15], [16]. Other work demonstrates vehicle-to-UAV communication or small-node aerial V2X feasibility, including emergency-response and detect-and-avoid use cases [17]. Berton et al.'s study is especially relevant: it investigates 3GPP C-V2X/PC5 for aerial collision avoidance and Broadcast Remote ID, and prototypes flight-path exchange using two drones and a ground control station equipped with C-V2X modems [18]. These works establish important aerial V2X feasibility results, but not the purely all-airborne topology evaluated here.

More directly related UAS-to-UAS and AAM communication studies identify direct aircraft-to-aircraft exchange as an enabler for cooperative collision avoidance, merging, separation assurance, right-of-way negotiation, and dynamic airspace coordination [19]–[27]. These studies establish that tactical aerial communication is a recognized need, but remain primarily use-case, demonstration, simulation, standardization, or requirements-level contributions. Other work studies sidelink, full-duplex, directional-antenna, air-to-air channel, and UAV broadcast architectures, often through simulation, analytical modeling, or channel characterization [28]–[34]. These efforts characterize connectivity, channel behavior, and candidate message pathways.

Across these categories, prior aerial V2V/A2X work generally retains a ground node, remains simulation- or requirements-based, characterizes channels without coordination semantics, or focuses on awareness, formation, or detect-and-avoid rather than tactical separation provision. We are not aware of prior work that couples a real all-airborne sidelink-class V2V stack to a tactical coordination controller and evaluates closed-loop outcomes under density, observability, integrity, and shared-resource stress.

Aerial V2V also differs from ground V2X in ways that limit direct transfer of ground C-V2X results: airborne links have three-dimensional LOS/NLOS behavior, UAS mobility and reaction horizons differ from lane-constrained traffic, and operational density is volumetric rather than lane-based. These differences make aerial tactical V2V a distinct neighborhood-exchange problem that must be characterized under aerial density, geometry, and controller-coupled stress.

### *C. Cooperative Perception and Coordination Semantics*

Cooperative perception work in ground V2X and emerging aerial/air-ground datasets shows that shared information can extend awareness beyond single-sensor range and through occlusion [35]–[43]. These works primarily evaluate perception accuracy, bandwidth-accuracy tradeoffs, or latency robustness; they do not evaluate whether shared information preserves the local belief state needed by an aerial tactical coordination controller, nor do they measure controller-coupled outcomes such as track drops, conflict events, reacquisition burden, or backstop activation.

Beyond cooperative perception, ground V2X message standards provide a useful protocol reference. ETSI CAM/DENM, SAE J2735 BSM/event messages, and IEEE 1609.2 security establish a periodic-plus-event-triggered communication pattern with authenticated messages [44]–[48]. This pattern informs our protocol structure, but our contribution is its adaptation to aerial tactical coordination, where exchanged content, freshness/integrity handling, and event-triggered coordination semantics are designed for controller-coupled local coordination and separation rather than ground-vehicle awareness or environmental notification alone.

### *D. Position of This Paper*

The literature leaves a specific gap at the intersection of UTM tactical separation, aerial V2V communication, and controller-coupled transportation outcomes. Strategic UTM is increasingly well defined, but tactical separation lacks a deployable evaluated neighborhood communication mechanism. Awareness links provide identification or state broadcast, but not the freshness, trust, intent, observability, and coordination semantics required for tactical separation. Aerial V2V/A2X work has established feasibility, standards direction, and collision-avoidance use cases, but has not characterized an all-airborne intent-first V2V neighborhood exchange stack coupled to closed-loop tactical coordination outcomes under high-volume stress. This paper fills that gap and follows the T-ITS tradition of evaluating communication through downstream transportation performance rather than through link metrics alone, including stability, merging, safety, perception quality, and flow [49]–[51]. We adopt that controller-coupled evaluation philosophy for low-altitude airspace.

## III. Intent-First V2V Coordination Protocol

This section defines the tactical V2V coordination protocol evaluated in the paper. The protocol is not a replacement for strategic UTM intent exchange, Remote ID, surveillance, or onboard collision avoidance. Its purpose is narrower: to support neighborhood-scale tactical coordination during short-horizon interactions.

### A. Protocol Design

The protocol uses two channels. A periodic state and intent beacon maintains the local belief state used for awareness continuity, cooperative perception, degraded-mode assessment, and predictive conflict evaluation. An event-triggered coordination channel is used only when aircraft need to negotiate, assert, cancel, or complete a local interaction outcome, such as yielding, admission, sequencing, release, rejoin, priority handling, or contingency behavior.

The protocol follows six design principles. Neighborhood scope: messages are intended for the local interaction set rather than city-wide real-time dissemination. Intent-first content: aircraft exchange short-horizon intent and maneuver authority, not only position, so neighbors can predict near-term behavior within bounded local authority. Freshness bounds: state, intent, and commitments expire and are not treated as authoritative after their validity window. Integrity-aware trust: messages carry authentication and integrity metadata so stale, spoofed, replayed, low-integrity, or otherwise invalid data can be rejected or down-weighted before control use. Observability continuity: periodic beacons carry awareness-quality, observability, and track-quality information to support cooperative perception and maintain local belief under degraded sensing. Coordination semantics: event-triggered messages express lifecycle state and tactical function, including proposal, commitment, abort, completion, yielding, admission, sequencing, release, rejoin, priority, contingency, and hazard-clear behavior.

These principles are what distinguish the proposed exchange from awareness-only broadcast. Awareness supports detection; intent-first tactical V2V supports a controller-usable local belief state, explicit local agreement, and safe degradation under uncertainty.

### B. Periodic Tactical Beacon

The periodic beacon provides the continuously refreshed local belief state used by the tactical controller. It carries identity, timing, sequence, and freshness metadata; kinematic state and short-horizon intent; awareness-quality and observability information; navigation-integrity and operating-state indicators; and bounded maneuver authority. The intent field is local and advisory rather than a globally authoritative route: it represents the aircraft's current short-horizon tactical plan only while freshness, integrity, and maneuver-authority conditions hold. The beacon therefore supports the controller-facing question: what can this aircraft safely assume about a neighbor over the next few seconds, how well observed is that neighbor, and how should reliance change if the information becomes stale, low-integrity, or degraded? Detailed beacon fields are provided in the supplementary material.

### C. Event-Driven Coordination Messages

Periodic beacons are insufficient when aircraft must coordinate explicit local actions. Dense tactical interactions require state transitions that cannot be inferred reliably from position alone, including yielding, admission, sequencing, release, rejoin, priority handling, and contingency behavior.

The event-triggered channel separates coordination lifecycle states from coordination functions. Lifecycle states define the transaction flow: PROPOSE initiates a bounded local request, COMMIT confirms execution within a validity window, ABORT cancels the interaction when assumptions fail, and CLEAR marks completion. Coordination functions define the tactical purpose: YIELD_PASS supports local yielding or passage order; ADMISSION, SEQUENCING, REJOIN, and RELEASE support shared-resource entry, ordering, reinsertion, and resource freeing; and PRIORITY, CONTINGENCY, and HAZARD_CLEAR support priority-aware, degraded, emergency, or hazard-zone behavior.

Messages are locally broadcast and then filtered by receivers based on relevance, participant set, zone reference, freshness, authority validity, and trust. This allows non-participant neighbors to observe nearby commitments when useful, without requiring every local interaction to be centrally arbitrated. The protocol is intentionally not a brittle handshake. If a proposal receives no response, the controller falls back to periodic intent and local state estimation. If a commitment message is lost, the receiver does not assume indefinite agreement. Commitments are short-lived, freshness-bounded, and backed by timeout logic, uncertainty inflation, degraded-mode behavior, and last-resort backstop activation if predicted margins collapse. The full message schema and coordination-primitive mapping are provided in the supplementary material.

### D. Freshness, Trust, and Degraded Operation

Freshness and trust rules determine whether received information may affect tactical control. Messages that fail authentication, replay, sequence, integrity, or expiry checks are rejected before control use. Messages that pass source validation but are stale, incomplete, low-confidence, or only partially covered by the local neighborhood may still inform prediction with reduced authority, but they trigger conservative behavior such as uncertainty inflation, reduced reliance on commitments, smaller trusted-neighbor sets, lower shared-resource admission capacity, and increased fallback or backstop readiness. This distinction between rejection and degradation is central: tactical V2V must specify not only how aircraft coordinate when information is good, but also how they stop relying on coordination when information quality deteriorates.

## IV. Evaluation Methodology

This section describes the field-anchored methodology used to evaluate the two-channel V2V protocol and its controller-coupled effects on tactical coordination and separation. The evaluation characterizes both the communication envelope under density and impairment and the resulting closed-loop outcomes, treating PRR and latency as inputs to transportation performance rather than standalone link metrics. Additional details on evaluation stack, scenario configurations, metric definitions, statistical reporting, and extended results are provided in the supplementary material.

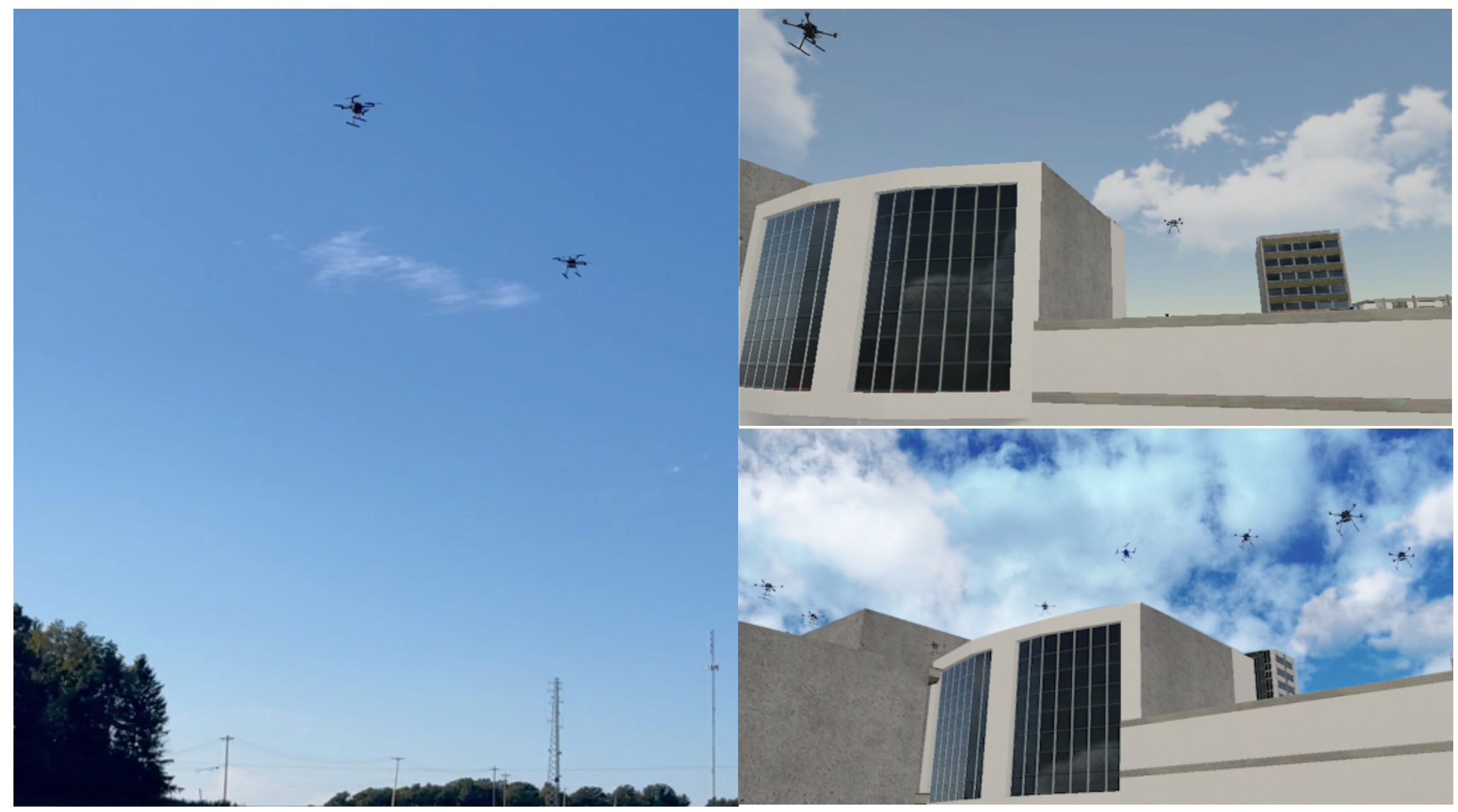

Fig. 1. Real-time field-anchored evaluation infrastructure. Left: physical multi-drone field operation. Upper right: corresponding field-synchronized real-time simulation view of the 1:1 urban operating environment. Lower right: hybrid operation coupling live hardware nodes with real-time simulated traffic, urban context, and scenario constraints. The stack exercises operational complexity around physical flight in real time, grounding communication and tactical-coordination results in closed-loop hardware behavior.

### *A. Real-Time Evaluation Stack and Evidence Layers*

The evaluation uses a real-time, field-anchored airspace stack that couples planning, communication, control, and local traffic interaction during execution (Fig. 1). The stack combines large-scale field-anchored simulation, hardware-, software-, and communications-in-the-loop evaluation, standalone field testing, and hybrid sim-field testing. These layers support two purposes: grounding the all-airborne V2V implementation in real hardware behavior, and extending the same controller and communication logic into density and impairment regimes that cannot be physically flown at scale. The evidence base spans 141,944 runs, 9,644 evaluation hours, and 74.3 million executed flight trajectories. In the hybrid layer, operational complexity is exercised around live flights in real time, allowing physical aircraft to interact with additional traffic, environment complexity, constraints, and communication effects that would not otherwise be present at the field site while preserving closed-loop behavior on actual hardware. The scaled simulation is then built on those field-anchored scenario families and operational behaviors, extending the same real-time evaluation logic into denser and more highly repeated regimes than can be executed physically.

The evaluated scenarios use two environment classes: synthetic custom urban environments from the real-time simulation pipeline, including 1:1-scale maps matched to the field environment, and city-anchor maps for New York City and Los Angeles. The synthetic environments support controlled real-time and hybrid sim-field testing, including field-site-aligned scenarios and repeatable disturbance conditions. The city-anchor maps are used in the large-scale field-anchored simulation layer to expose the same controller and communication logic to realistic urban geometry, hotspot structure, and LOS/NLOS obstruction patterns. Density is reported as projected areal density over the active operational footprint rather than as lane-based traffic density; vertical structure is represented through scenario geometry, altitude bands, and local interaction rules. Evaluated stress conditions span densities up to 250 vehicles/km², with the upper end used to characterize communication-control stress and failure-boundary behavior rather than routine operating capacity. Unless otherwise stated, the evaluated traffic uses a homogeneous managed UAS class so that differences across baselines reflect coordination, communication, and degraded-mode behavior rather than mixed vehicle-performance effects. Full evidence-layer details are provided in the supplementary material.

### *B. All-airborne V2V Hardware Implementation*

The evaluated communication stack uses sidelink-class C-V2X/PC5 modules configured for direct air-to-air exchange. The direct hardware configuration used six airborne V2V nodes with no ground relay, mounted on mid-size UAS platforms (Fig. 2) carrying the V2V module, antenna, and GNSS. GNSS was also used as the NTP reference for application-layer timestamp synchronization. Periodic intent beacons are transmitted at 10 Hz using a 100 ms semi-persistent scheduling period, while event-triggered coordination messages are generated through a separate event-flow path only when local interactions require explicit coordination. The evaluated profile used a 5.9 GHz ITS-band carrier as an experimental carrier for the studied tactical-neighborhood exchange. This should not be interpreted as a claim that the band is a sufficient long-term aviation spectrum allocation. Additional hardware-integration details are provided in the supplementary material.

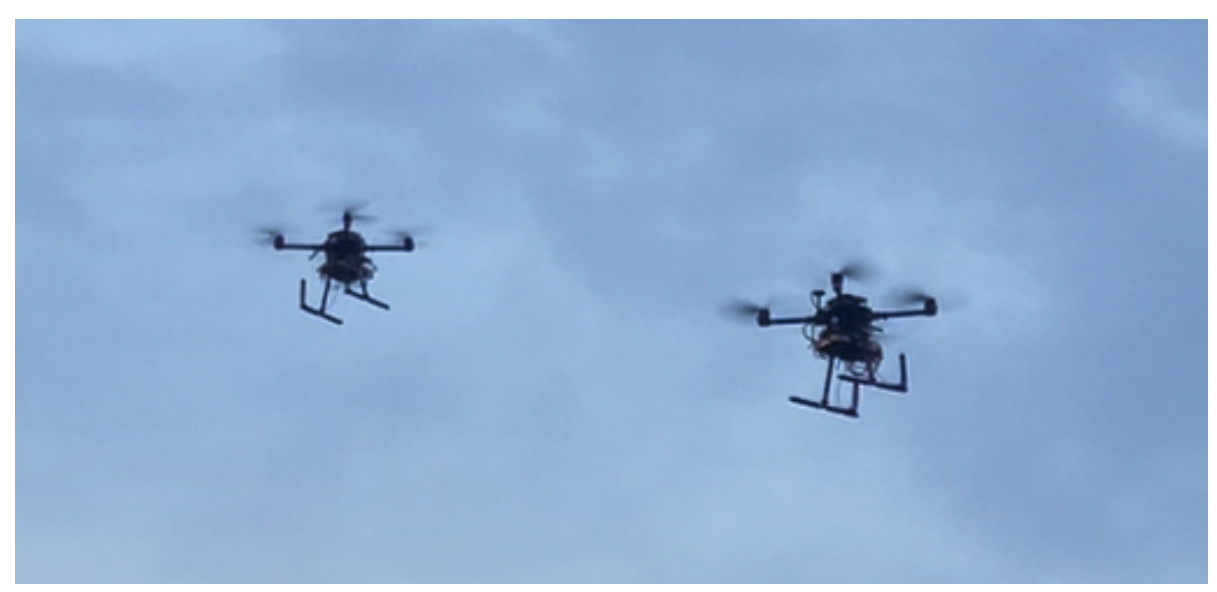

Fig. 2. Airborne V2V hardware integration. Field drones carrying V2V modules and antennas for direct air-to-air neighborhood exchange.

### C. Tactical Controller and Experimental Baselines

The tactical controller is the closed-loop substrate through which communication effects become operationally measurable. The paper does not introduce the controller as a standalone planning contribution. The evaluation uses three baselines, all with the same onboard last-resort collision-avoidance backstop. Baseline A, strategic-only, performs conflict management through pre-flight intent coordination and in-flight reauthorization using a Discovery and Synchronization Service (DSS), consistent with UTM operational evaluations. DSS is a UTM function to discover and synchronize strategic operational intent and constraints. Baseline A does not include a dedicated tactical coordination controller. Baseline B1, non-V2V tactical, uses the tactical coordination controller without the V2V communication channel. Baseline B2, V2V-enabled tactical, uses the same tactical coordination controller with authenticated, freshness-bounded neighborhood exchange through V2V, including cooperative perception, local sequencing, and guarded degraded-mode behavior.

This baseline structure supports two comparisons. A versus B2 tests whether deployable V2V-enabled tactical coordination improves over strategic-only coordination under disturbance and density. B1 versus B2 isolates both the operational role of V2V-enabled neighborhood exchange and the cost of making that exchange deployable under authentication, bounded freshness, congestion, coverage limits, and degraded-mode behavior.

Both tactical baselines use the same hybrid predictive local controller family, including short-horizon conflict prediction, yielding/commitment logic, uncertainty inflation for stale or delayed intent, vertical separation handling, shared-resource admission behavior, and onboard backstop activation. Backstop behavior activates when predicted separation or time-to-conflict falls below the scenario-defined trigger thresholds.

### D. Communication, Impairment, and Propagation Modeling

Communication stress is modeled through two distinct impairment families. The first family perturbs the tactical V2V exchange itself by varying one-way latency, jitter, and packet loss, thereby affecting message freshness, deadline misses, and neighborhood consistency (Table I). The second family perturbs the propagation of dynamic operational context, such as temporary obstacles, changed constraints, or updated airspace-validity information, thereby affecting when aircraft become aware of changed conditions (Table II). Separating these families allows the evaluation to distinguish failures caused by degraded neighborhood communication from failures caused by delayed or incomplete context availability.

TABLE I
Neighborhood Communications Impairment Classes

| Class | Level | One-way latency | Jitter | Packet loss |
|---|---|---|---|---|
| **N0** | Nominal | 20 ms | 5 ms | 0.5% |
| **N1** | Mild | 50 ms | 15 ms | 1% |
| **N2** | Moderate | 120 ms | 30 ms | 3% |
| **N3** | Severe | 250 ms | 50 ms | 6% |

TABLE II
Context-Update Impairment Classes

| Class | Impairment level | Radius | Propagation Delay | Partial Delay | Propagation Completeness | Local Relay Delay |
|---|---|---|---|---|---|---|
| **C0** | nominal | 140 m | 2.0 s | 1.0 s | 1.0 | 0.5 s |
| **C1** | mild | 160 m | 6.0 s | 2.0 s | 0.9 | 0.75 s |
| **C2** | moderate | 180 m | 14.0 s | 5.0 s | 0.7 | 1.0 s |
| **C3** | severe | 200 m | 24.0 s | 8.0 s | 0.5 | 1.25 s |

Urban propagation is modeled using a hardware-informed geometry-aware stress model rather than site-calibrated RF ray tracing. Baseline links use configured latency, jitter, packet loss, queueing, and PRR-versus-distance behavior informed by hardware and hybrid evaluation layers. Selected city-anchor sweeps compute LOS/NLOS state from building-volume intersections and apply NLOS range, packet-loss, and latency penalties, with material-aware attenuation where map-derived building classes are available.

### E. Metrics

Table III summarizes the principal metrics used in the main paper; full definitions and scenario-specific qualification rules are provided in the supplementary material.

TABLE III
Metric Categories and Purpose

| Metric category | Metrics | Purpose |
|---|---|---|
| **Communication** | PRR, latency (p95), deadline misses, information age | Characterize timing, freshness, and load behavior of the V2V exchange |
| **Integrity/trust** | invalid message rejections, bad accepts, false-clearance/conflict events | Measure trust enforcement and false inference under invalid or low-integrity information |
| **Operational performance** | throughput, safety-qualified throughput, mean hold time, hold burden | Measure traffic continuity and operational usability |
| **Strategic burden** | replans, replans per flight, DSS/query burden where applicable | Measure burden shifted into strategic reauthorization |
| **Safety** | minimum separation, conflict events, conflict steps, near misses | Measure separation maintenance and conflict persistence |
| **Stability** | deadlock, oscillation burden, blast radius | Measure local instability and disturbance propagation |
| **Perception/ observability** | track drops, reacquisition time, track continuity | Measure observability continuity under degraded sensing |
| **Terminal/ shared-resource behavior** | wave-offs, queue peak, recovery time, release/rejoin behavior | Measure hotspot and terminal sequencing stability |
| **Backstop behavior** | backstop activations, activation duration | Measure reliance on last-resort safety behavior |

In the results, throughput denotes completed operations per hour; minimum separation is the smallest observed pairwise aircraft separation; conflict events are discrete conflict episodes; track drops are losses of previously usable local tracks; and safety-qualified throughput counts completed operations only for windows satisfying scenario-specific safety and stability criteria. For principal baseline comparisons, paired 95% confidence intervals were computed over matched repeated runs or scenario-condition summaries, with pairing by scenario family, density, stress class, and random seed where applicable. Full aggregation details are provided in the supplementary material.

### *F. Scenario Suite*

The evaluation uses a scenario suite designed to expose how intent-first V2V affects tactical coordination and separation under density, disturbance, degraded information, and shared-resource stress. Scenario families are evaluated over scenario-specific density and stress sweeps rather than a single uniform grid. Across the main analyses, the density range covers the lower end of the evaluated stress range at 50 vehicles/km², intermediate transition regimes at 100–120 vehicles/km², dense degradation regimes at 150–200 vehicles/km², and diagnostic breakdown stress up to 250 vehicles/km² where applicable. These densities are intentionally higher than those used in prior UTM flight demonstrations [26], [27] and are used to stress-test the evaluated communication-control envelope for future scaled operations, not as claims of current routine operating density. Each scenario is initialized at the target density and baseline configuration, after which the relevant stressor is injected according to the scenario-specific parameterization. Outputs are computed over the full run or over the active disturbance/validity window depending on the metric, and baselines are compared under matched density, stress, and seed conditions. Table IV summarizes the scenario families and the primary V2V mechanism each family tests. Detailed scenario grids, qualification criteria, and extended outputs are provided in the supplementary material.

TABLE IV
Scenario Families

| Scenario family | Primary mechanism tested |
| --- | --- |
| **Communications impairment** | Stale-belief divergence under latency, jitter, packet loss, and reduced freshness |
| **Dynamic context-update latency** | Stale-validity window during delayed or incomplete propagation of changed airspace context |
| **Tactical message-integrity stress** | Authentication, replay/spoof rejection, and trusted coordination |
| **Navigation-integrity/GNSS corruption** | False-inference suppression under corrupted or low-integrity state |
| **Degraded observability** | Cooperative-perception support and track-continuity preservation |
| **Hotspot / pad jitter / wave-off** | Local admission, sequencing, release, and rejoin behavior around shared resources |
| **Multi-intruder burst near hotspot** | Compound local coordination under burst disturbance and shared-resource stress |
| **Mixed equipage / intruder injection** | Partial-participation coordination among cooperative aircraft under uncooperative traffic |

The baseline comparisons reported in Section VI are selected to match the mechanism being evaluated. Communication-impairment and dynamic context-update scenarios focus on A-B2 comparisons because they test whether V2V-enabled tactical coordination changes the degradation mode relative to strategic-only coordination when neighborhood information or changed context becomes delayed, incomplete, or stale. Message-integrity stress compares authenticated B2 with an authentication-disabled B2 ablation because the stressor acts on the V2V message-validation path. Integrity and degraded-observability scenarios focus on B1-B2 comparisons because the non-V2V tactical baseline isolates tactical behavior without V2V-supported trust, freshness, observability, or degraded-mode constraints. Shared-resource sequencing focuses on A-B2 comparisons because admission, sequencing, release, and rejoin require explicit shared coordination state. Mixed-equipage scenarios report A, B1, and B2 because all three baselines represent distinct operational responses to uncooperative traffic.

## V. Communication Envelope Under Density and Impairment

This section characterizes the communication envelope of the evaluated all-airborne intent-first V2V exchange before examining controller-coupled outcomes in Section VI. The objective is to determine when the neighborhood exchange remains fresh enough to support tactical coordination, and when it transitions into a stale-belief regime.

### *A. PRR and Latency Envelope*

The primary communication metrics are Packet Reception Ratio (PRR), 95th-percentile latency, and deadline-miss rate. PRR measures effective neighborhood reception under continuous multi-node broadcast, not isolated single-link performance. This distinction matters because tactical coordination depends on whether the local neighborhood maintains sufficiently fresh and consistent information under load, not whether a single aircraft pair can communicate under favorable conditions. Latency is measured from application-layer send timestamp to receiver-side message delivery availability. Deadline misses indicate messages that arrive too late for tactical use under the fixed 250 ms end-to-end deadline.

Fig. 3a shows the effective neighborhood PRR across density and impairment class. PRR remains strong at 50 vehicles/km², 0.820–0.853, and usable at 100 vehicles/km², 0.708–0.747. It falls to a degraded regime at 150 vehicles/km², 0.549–0.595, and weakens further at 250 vehicles/km², 0.310–0.314. Under severe impairment, PRR drops from 0.820 at 50 vehicles/km², 95% CI [0.814, 0.827], to 0.549 at 150 vehicles/km², 95% CI [0.515, 0.583], and 0.310 at 250 vehicles/km², 95% CI [0.290, 0.331]. This shows a density-driven transition from strong to degraded and then weak neighborhood exchange.

Fig. 3b shows the corresponding latency (p95) envelope. Under nominal impairment, latency (p95) rises from 88.5 ms at 50 vehicles/km², 95% CI [77.2, 99.7], to 130.2 ms at 150 vehicles/km², 95% CI [129.2, 131.3], and remains near that level at 250 vehicles/km², 131.4 ms, 95% CI [130.9, 131.9]. Under severe impairment, latency (p95) is already 333.4 ms at 50 vehicles/km², 95% CI [323.8, 343.0], and rises to 390.4 ms at 250 vehicles/km², 95% CI [389.9, 391.0]. Thus, density primarily erodes PRR, while severe impairment strongly inflates latency; together, they determine whether neighborhood state and intent remain usable for tactical coordination.

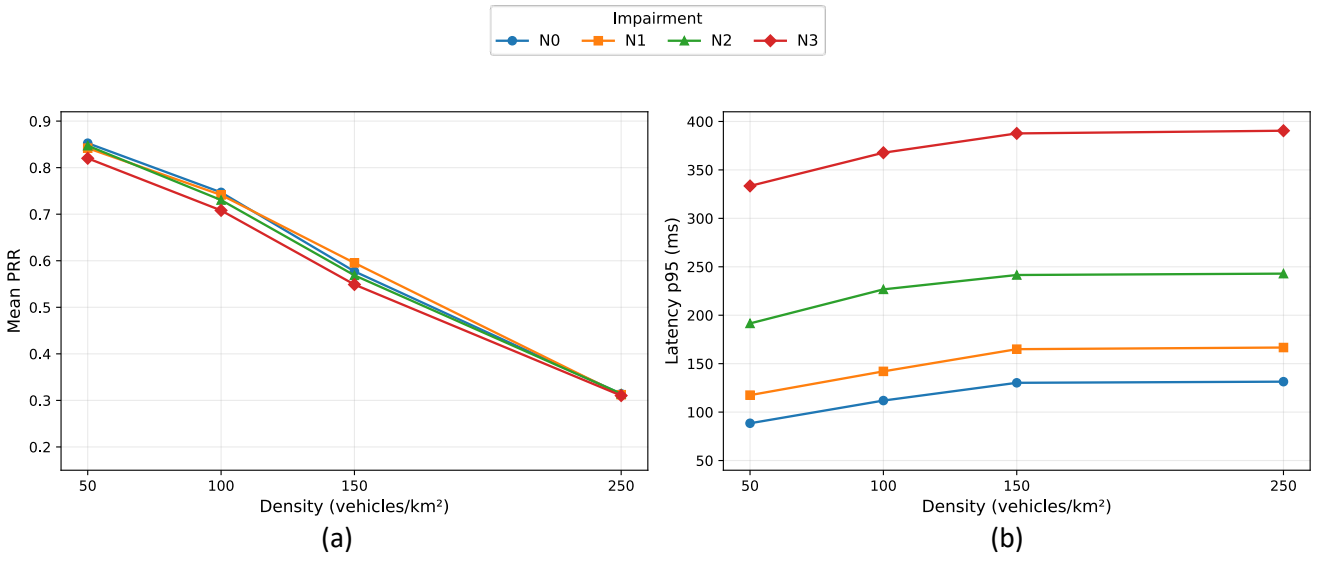

Fig. 3. (a) Neighborhood PRR and (b) latency p95 across density and impairment class. PRR is measured as effective neighborhood reception under continuous multi-node broadcast. The 250 vehicles/km² slice is included as a diagnostic breakdown stress point.

Deadline-miss behavior shows the transition more sharply. Deadline misses remain effectively zero under nominal and mild impairment through 250 vehicles/km² and remain near zero under moderate impairment, rising only to 0.0055 at 250 vehicles/km², 95% CI [0.0052, 0.0058]. Under severe impairment, however, the deadline-miss rate reaches 0.756 at 50 vehicles/km², 95% CI [0.717, 0.796], 0.906 at 150 vehicles/km², 95% CI [0.904, 0.907], and 0.906 at 250 vehicles/km², 95% CI [0.902, 0.909]. Severe impairment therefore changes the operating mode by making a large fraction of neighborhood messages untimely for tactical use.

The communication envelope is also range- and geometry-dependent. In the broader communication characterization, representative effective neighborhood ranges span roughly 700–1300 m depending on density, geometry, and propagation condition, including LOS and urban NLOS slices. At those operating points, effective PRR remains around 0.70–0.73 with latency (p95) around 100–107 ms. These values should be interpreted as effective neighborhood performance under the evaluated tactical profile, not as single-link maximum radio range.

### *B. Density Transition and Stale-Belief Regime*

The communication results define three regimes. In the lower-density regime of the evaluated stress envelope, around 50 vehicles/km², the exchange remains sufficiently reliable for tactical coordination under nominal and moderate impairment. In the intermediate regime, around 100-150 vehicles/km², degraded PRR and increased latency make consistently fresh neighbor intent unreliable, requiring communication-aware degraded behavior. In the highest-density stress regime, represented by 250 vehicles/km², the exchange becomes dominated by missed updates, stale information, and insufficient neighborhood consistency.

This transition is important because tactical coordination can fail even when connectivity has not disappeared entirely. A message that arrives late, inconsistently, or without sufficient neighborhood coverage can still produce an outdated local belief. The communication problem therefore shifts from "is there a link?" to "is the local information fresh and consistent enough for control?"

### *C. Operational Interpretation of the Communication Envelope*

The key result of this section is the shape of the transition rather than a single maximum density or range. The evaluated V2V stack supports tactical coordination in lower-density neighborhoods, becomes increasingly scenario-dependent in intermediate regimes, and enters a stale-belief and degraded regime at the highest stress levels. The 250 vehicles/km² point should be treated as a diagnostic breakdown endpoint, not an operating regime for the evaluated implementation.

This envelope explains why the controller-coupled results in Section VI are mechanism-specific. V2V can reduce stale belief, support cooperative perception, enable trusted coordination, and structure local sequencing only while the neighborhood exchange remains sufficiently fresh and reliable for tactical use. As density and impairment increase, the controller must shift from cooperative tactical behavior toward uncertainty inflation, reduced trusted-neighbor sets, guarded fallback, and backstop readiness.

## VI. Controller-Coupled Operational Results

Section V characterized the communication envelope of the evaluated V2V stack. This section evaluates how that envelope affects closed-loop tactical coordination and separation outcomes. Results are organized by mechanism rather than by scenario number to show how different stressors expose different operational roles for V2V-enabled tactical coordination.

### *A. Stale-belief Reduction Under Communication Impairment and Context Delay*

Communication impairment and delayed context propagation test whether tactical coordination remains useful when local information arrives late, incompletely, or inconsistently. Under communication impairment, the strategic-only baseline absorbs degradation as increased coordination burden. Across the communications-impairment comparison at 150 vehicles/km², increasing impairment from nominal to severe lowers strategic-only throughput by 312 operations/hr, from 1,152/hr to 840/hr, while DSS query load increases by 334 per run, from 12,921 to 13,255, and mean hold time remains high at roughly 135–138 s.

The V2V-enabled tactical controller changes the degradation mode by localizing response at the neighborhood level rather than pushing the burden into strategic reauthorization. At 150 vehicles/km², increasing impairment from nominal to severe reduces B2 throughput from 853/hr to 507/hr, a similar directional throughput loss to the strategic-only baseline, but eliminates the hold burden observed under strategic-only coordination. Instead, degradation appears as increased local fallback reliance: backstop activations increase from 682 under nominal impairment to 717 under severe impairment, with activation duration (p95) on the order of 45-50 s. Thus, V2V does not eliminate degradation; it shifts the response from global coordination burden and holding toward local guarded operation and backstop collision avoidance.

Delayed context propagation produces a related stale-validity problem. In the dynamic context-update comparison, the V2V-enabled tactical controller increases throughput relative to the strategic baseline by 486 operations/hr, 95% CI [266, 706], receives and reacts to updated context 3.42 s earlier, 95% CI [2.60, 4.23], and reduces window holds by 97.5, 95% CI [81.0, 114.0]. At 50 vehicles/km², this corresponds to V2V-enabled tactical controller sustaining 1960–2660 operations/hr with no window holds across context-delay classes, compared with 1060–1290 operations/hr and 28–79 window holds for the strategic baseline.

The relevant metric is not raw convergence time alone: strategic-only coordination can appear to converge by immobilizing traffic, while local tactical coordination attempts to preserve flow and restabilize the neighborhood. Thus, V2V reduces stale-belief divergence while the neighborhood retains sufficient freshness and coverage; beyond that envelope, the controller shifts toward guarded fallback.

### *B. Trusted Coordination Under Message-Integrity Faults*

Message-integrity stress tests evaluate spoofing, replay, reordering, duplication, and mixed message faults. This mechanism is distinct from packet loss: a lost message removes information, while an accepted invalid message can create a false commitment, false clearance, or incorrect sequencing state.

Fig. 4 isolates this trusted-coordination mechanism. With authentication enabled, invalid tactical messages are rejected before entering the control loop. With authentication disabled, the controller accepts large volumes of invalid content under the same fault conditions. In the spoof-fault case at 50 vehicles/km², authenticated V2V rejects 166,744 invalid messages with zero bad accepts, while the unauthenticated variant accepts 163,874 bad messages. In the replay-fault case at the same density, authenticated V2V rejects 161,082 invalid messages with zero bad accepts, while the unauthenticated variant accepts 159,170 bad messages. In the heavier mixed-fault case at 150 vehicles/km², authenticated V2V rejects 5,168,205 invalid messages with zero bad accepts, while the unauthenticated variant accepts 5,349,010 bad messages.

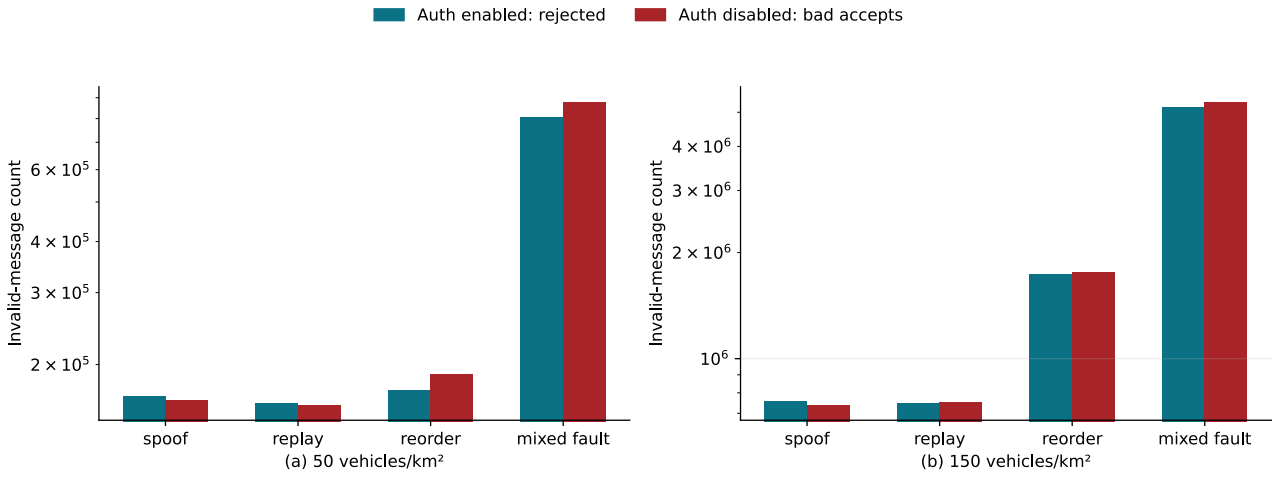

Fig. 4. Tactical message-integrity handling with authentication enabled and disabled. Panels compare invalid-message rejections under authenticated B2 with bad accepts under the authentication-disabled B2 ablation across representative fault classes at 50 and 150 vehicles/km². Counts are shown on a log scale.

This result establishes authentication as a tactical safety mechanism. Because V2V messages affect yielding, commitment, ordering, release, and contingency behavior, untrusted messages can alter the controller's local decision state. Trusted exchange is therefore a separation-supporting property, not merely a cybersecurity add-on.

### *C. False-Inference Suppression Under Corrupted or Low-Integrity State*

Integrity stress also exposes a second mechanism: false-inference suppression. The hazard is not only that a bad message is accepted, but that the controller forms a false local belief and acts on it. GNSS spoofing, low-integrity state, stale intent, and replayed coordination can generate false conflicts, false clearances, or unsafe commitments if treated as valid tactical inputs.

Under high-severity spoofing at 200 vehicles/km², the strategic-only baseline records 16,096 false-clearance events and 9,970 false-conflict events. The non-V2V tactical baseline preserves aggressive local repair, but without deployable trust and degraded-mode constraints it can amplify false inference; in the same condition, it records 44,551 false-clearance events and 48,704 false-conflict events.

The V2V-enabled tactical controller reduces the propagation of false local belief through authentication, freshness checks, integrity-aware state handling, uncertainty inflation, and guarded fallback. At 50 vehicles/km² under high-severity spoofing, it sustains usable operation while reducing false-inference growth relative to the non-V2V tactical baseline. At higher densities, the controller becomes more conservative and eventually highly degraded, which marks the boundary where corrupted state and density exceed the deployable envelope. This shows that high-throughput tactical behavior without integrity-aware constraints can be unsafe; deployable V2V suppresses false inference by treating trust and state integrity as control-relevant variables.

Fig. 5 compares false-clearance and false-conflict events for the non-V2V tactical baseline and the V2V-enabled tactical controller under GNSS integrity stress. Across matched GNSS-integrity stress conditions up to 200 vehicles/km², V2V-enabled tactical controller reduces false-clearance events relative to non-V2V tactical baseline by 8478, 95% CI [4650, 12,305], and false-conflict events by 9738, 95% CI [5334, 14,143]. The result is that deployable V2V suppresses false tactical inference by refusing to act aggressively on corrupted state.

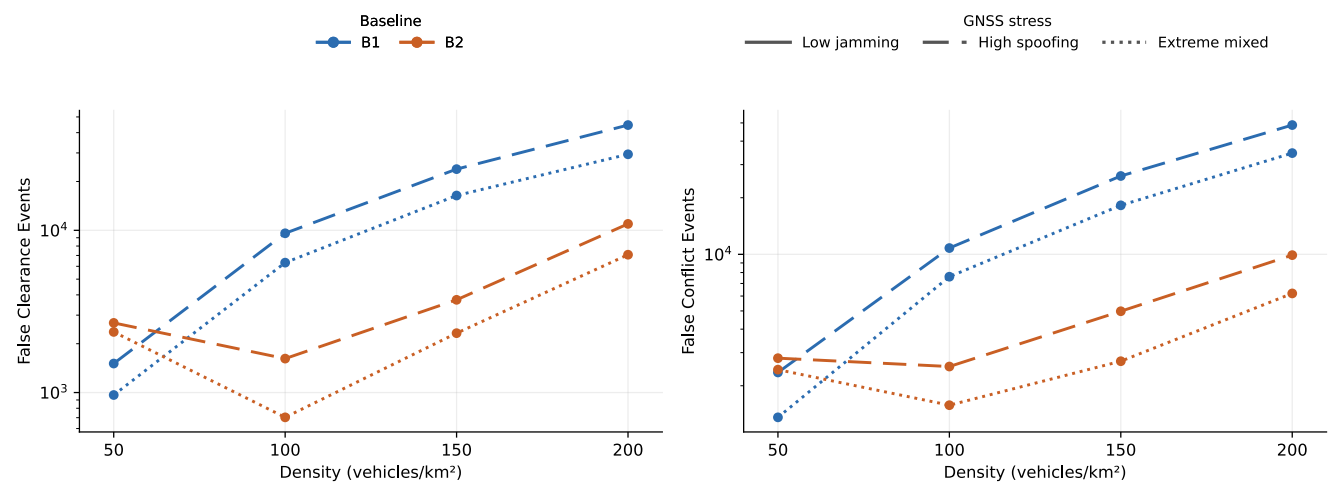

Fig. 5. False tactical inference under GNSS integrity stress. Panels compare false-clearance and false-conflict events for B1 and B2 across matched integrity-stress conditions.

### *D. Cooperative Perception Under Degraded Observability*

Degraded-observability scenarios evaluate whether V2V preserves the local belief state when onboard sensing becomes intermittent, occluded, or unreliable. The key outcome is track continuity, because tactical coordination depends on maintaining usable local belief under partial observability.

Fig. 6 isolates this cooperative-perception mechanism. Without V2V-supported cooperative perception, the non-V2V tactical baseline preserves raw flow by continuing local repair under weaker belief continuity, but this comes with high track-loss and conflict-event burden. With V2V, track drops fall by more than an order of magnitude across the density sweep. Across the degraded-observability comparison, the V2V-enabled tactical baseline reduces track-drop events relative to the non-V2V tactical baseline by 11,454 events on average, 95% CI [8,157, 14,751], and reduces conflict events by 15.8, 95% CI [9.6, 22.0]. At 150 vehicles/km², this corresponds to a reduction in track drops from 14,636 to 201 and in conflict events from 21.33 to 1.00.

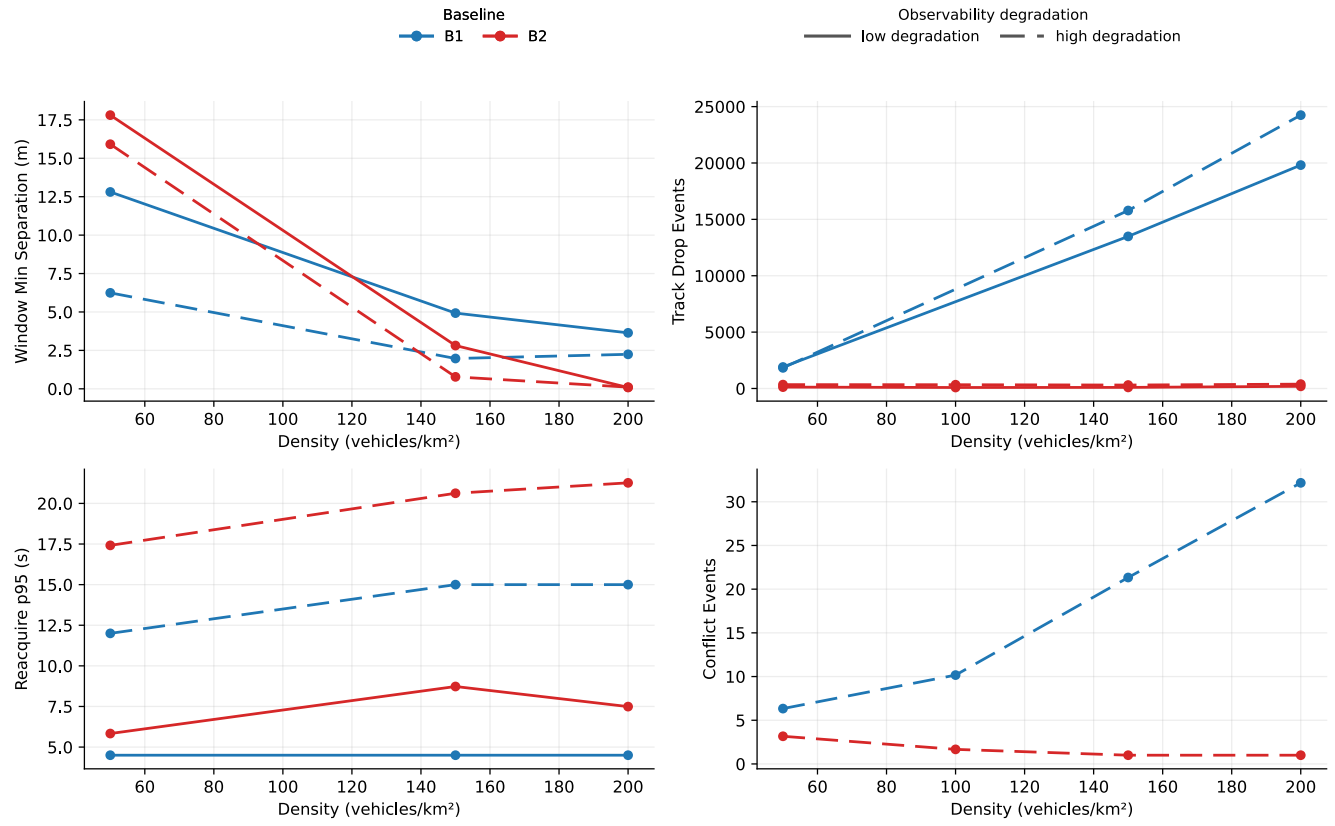

Fig. 6. Cooperative-perception effects under degraded observability. Panels compare window minimum separation, track-drop events, reacquisition time, throughput, and conflict events for B1 and B2 across the density sweep.

Thus, cooperative perception acts as belief continuity for the tactical controller, not simply as an auxiliary sensing function. The benefit remains bounded: in dense degraded slices, window minimum separation and reacquisition burden still degrade, and the V2V-enabled tactical controller sacrifices throughput as part of its conservative deployable behavior.

## *E. Structured Sequencing Around Shared Resources*

Hotspot and terminal scenarios test whether V2V supports local traffic-management behavior beyond pairwise avoidance. These scenarios include pad jitter, wave-offs, queue instability, and burst disturbances near shared resources. The relevant problem is admission, sequencing, release, retry, and rejoin around a constrained local resource.

Fig. 7 shows the pad jitter and wave-off scenario. Under strategic-only coordination, high nominal flow can coexist with unstable local behavior. In the high-disruption pad jitter case at 50 vehicles/km², the V2V-enabled tactical controller reduces queue peak by 13, from 17 to 4, and reduces wave-offs by 26, from 29 to 3, while improving the minimum-separation by 12.3 m, from 0.65 m to 12.92 m. This shows that event-driven coordination semantics can stabilize lower-density shared-resource interactions.

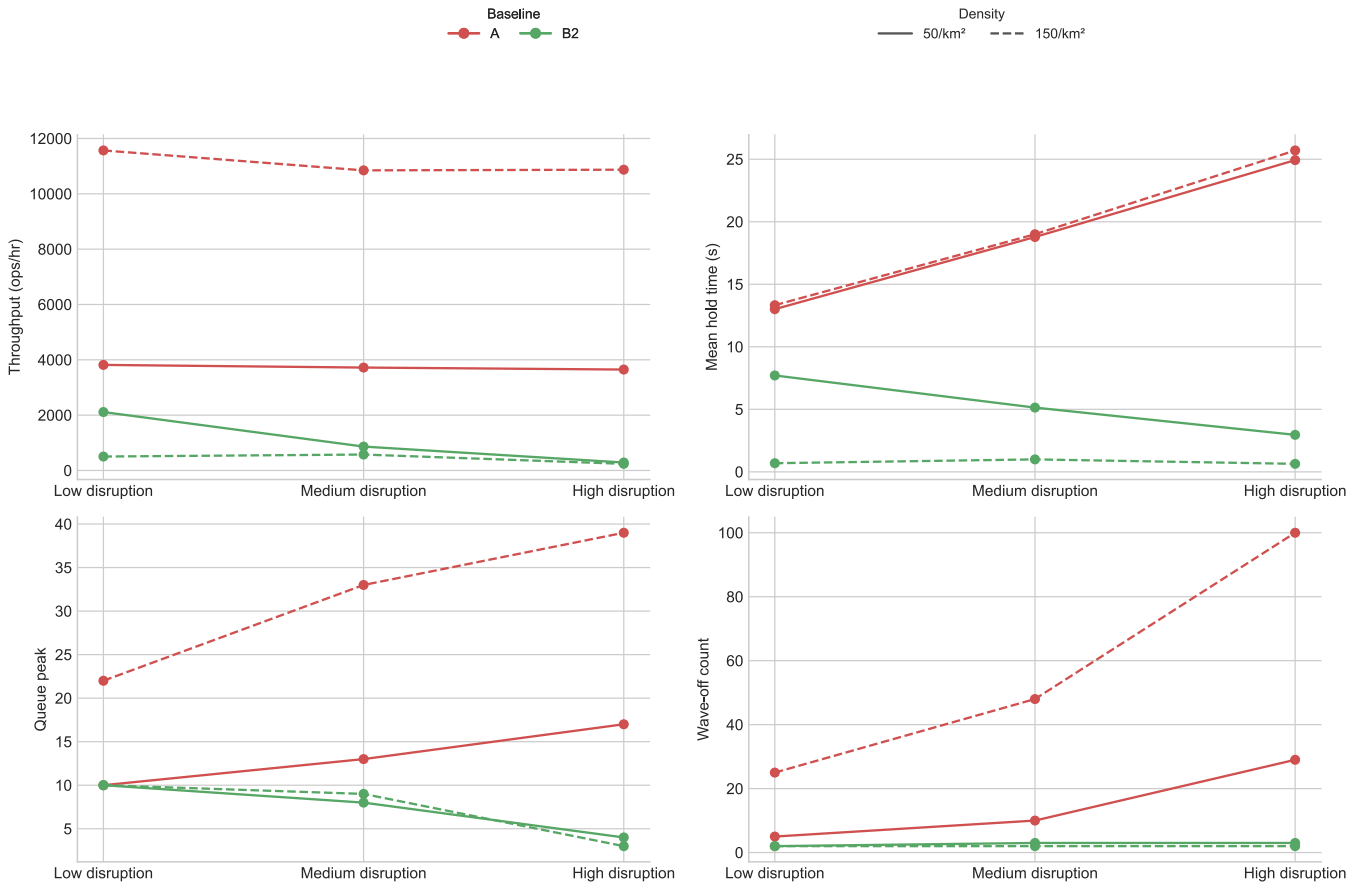

Fig. 7. Shared-resource sequencing under pad-jitter and wave-off stress. Panels compare throughput, mean hold time, queue peak, and wave-offs for A and B2 baselines across disruption levels and density conditions.

The benefit does not extend unboundedly. At 150 vehicles/km², throughput remains limited, minimum separation falls to 0.45–0.71 m across disruption levels, and deadlock occurrence rises substantially. Multi-intruder burst tests near the hotspot show the same pattern: V2V improves lower-density burst handling but does not fully resolve dense compound disturbances. Thus, event-triggered V2V coordination stabilizes lower-density shared-resource interactions, but dense hotspot management still requires stronger admission and sequencing logic.

## *F. Mixed-Equipage and Partial-Participation Coordination*

Mixed-equipage scenarios evaluate V2V behavior when not all aircraft participate in tactical exchange. V2V does not eliminate uncertainty from uncooperative traffic. Its value is conditional: it improves consistency among participating aircraft so they do not generate secondary conflicts while responding to uncertain encounters.

In a high intruder-disruption case at 50 vehicles/km², the strategic-only baseline records 1436 intruder detections and 42 intruder-conflict events, with throughput around 840 operations/hr. The non-V2V tactical baseline increases detections to 2483 but reduces intruder-conflict events to 6, showing that local tactical response improves conflict absorption. The V2V-enabled tactical controller eliminates intruder-conflict events in the tested lower-density case while maintaining throughput around 2240 operations/hr with mean minimum separation 15.6 m.

At higher densities, throughput and separation tradeoffs re-emerge. V2V improves coordination among cooperative participants, but onboard sensing, prediction, conservative margins, and backstop behavior remain necessary when shared intent is unavailable.

## *G. Aggregate Safety and Flow Tradeoffs*

The mechanism-specific effects also appear at the aggregate system level. Fig. 8 summarizes strategic-only coordination and the V2V-enabled tactical controller across the eight scenario families introduced in Section IV-F. For each density and baseline, results are aggregated over the evaluated disruption-severity sweep within each scenario family, with equal weighting across scenario families. Each density group includes 19,692 runs. Safety-qualified throughput denotes completed operations per hour counted only for runs or windows satisfying the applicable scenario-specific safety criteria, including no unresolved conflict persistence, no sustained separation violation, and no unrecovered backstop or deadlock condition. Qualified fraction denotes the fraction of evaluated runs or windows satisfying those criteria.

This aggregate comparison asks whether intent-first V2V extends usable operation beyond the strategic layer that would otherwise absorb disturbances through reauthorization, holding, or replanning. The non-V2V tactical baseline is used in the preceding mechanism-specific comparisons to isolate the role and cost of V2V; here, the comparison shows when tactical V2V becomes operationally necessary relative to strategic-only coordination.

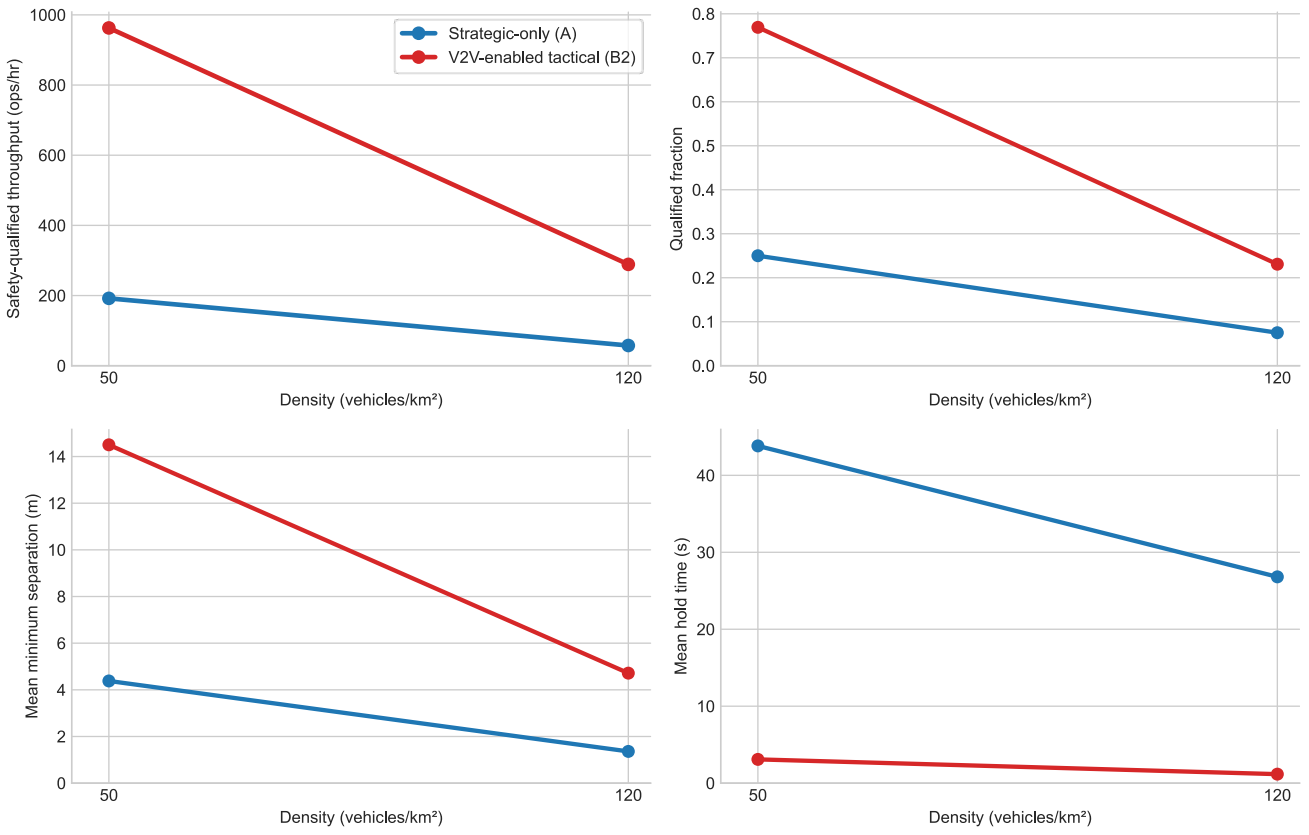

Fig. 8. Aggregate safety-qualified throughput and safety/flow tradeoffs across representative stressed scenarios. Panels report safety-qualified throughput, qualified fraction, mean minimum separation, and mean hold time for A and B2 across density conditions.

Across the lower-to-moderate stressed density regime, the V2V-enabled tactical controller retains higher safety-qualified throughput, higher qualified fraction, stronger mean minimum separation, and substantially lower hold burden. The degradation trend remains visible as density increases, indicating that V2V extends but does not remove the deployability boundary.

## VII. Discussion and Limitations

The results support three conclusions. First, tactical V2V value is mechanism-specific rather than a scalar improvement in link quality, throughput, or safety. Second, V2V becomes operationally important in local, high-coupling, post-dispatch regimes where strategic coordination would otherwise absorb disturbance through holding, replanning, or stale global state. Third, deployable V2V-enabled tactical coordination remains bounded by density, communication freshness, observability, trust, and local interaction complexity. In this sense, the evaluated V2V-enabled tactical layer is not a replacement for strategic coordination or a raw-throughput maximizer, but a deployable communication layer that allows tactical separation to become operationally useful when density, disturbance, or stale context exceed what strategic reauthorization alone can absorb.

### *A. V2V As Tactical Coordination Service*

The results show that the evaluated intent-first V2V is best understood as a bounded tactical coordination service rather than an awareness-only communication channel. Under communication impairment and delayed context propagation, V2V shortens the path by which local state, short-horizon intent, and changed operational context become usable to affected aircraft, reducing stale-belief divergence before disturbances spread into broader strategic holding or reauthorization. Under degraded observability, V2V acts as a belief-continuity mechanism: cooperative perception through periodic beacons helps preserve the local tracks needed for tactical prediction when onboard sensing becomes intermittent, occluded, or unreliable. Under message-integrity and navigation-integrity stress, V2V acts as a trusted-coordination mechanism, because authentication, freshness, and integrity state determine whether information can enter the control loop and how strongly the controller may rely on it. Around hotspots and shared resources, V2V acts as a local traffic-management mechanism: event-triggered messages provide the semantics needed for admission, sequencing, release, rejoin, priority, and contingency behavior, allowing constrained-resource interactions to be handled locally rather than pushed immediately into global reauthorization or broad holding. In mixed-equipage encounters, V2V does not remove uncertainty from uncooperative traffic, but it improves consistency among participating aircraft so they do not create secondary conflicts while responding to uncertain intrusions. Across these cases, the common role of V2V is to preserve usable neighborhood coordination while information remains fresh, trusted, and actionable, and to trigger conservative degradation when that assumption no longer holds.

### *B. Derived Communication Requirements for Aerial V2V*

The evaluation identifies six implementation requirements for tactical aerial V2V. First, state, intent, commitment, admission, sequencing, release, and contingency messages require bounded freshness: expired information should not remain authoritative for control. Second, authentication and integrity protection must occur before control use, because invalid messages can alter yielding, commitment, release, or sequencing state. Third, tactical exchange requires intent and coordination semantics beyond position broadcast, including who has yielded, who has committed, which aircraft is admitted to a shared resource, when an interaction is released, and how an aircraft should rejoin after a wave-off. Fourth, observability and track-quality information should be part of the tactical exchange because belief continuity under degraded sensing directly affects conflict growth and backstop reliance. Fifth, controllers must include communication-aware degraded operation, including uncertainty inflation, reduced reliance on negotiated commitments, smaller trusted-neighbor sets, lower admission capacity, and increased readiness for fallback or backstop behavior. Sixth, tactical V2V requires congestion-aware prioritization, because periodic state and intent beacons, commitment messages, releases, rejoin messages, and contingency signals do not have the same urgency under channel load. These requirements should be validated against the intended density, geometry, disturbance environment, and controller design. The observed PRR, latency, and density breakpoints are operating-envelope findings for the evaluated stack, not universal thresholds.

*C. Deployability Envelope and Failure Boundary*

The evaluated V2V-enabled tactical stack has a bounded but operationally meaningful communication-control envelope. Cooperative tactical use was strongest when effective neighborhood PRR remained high and latency p95 was near the 100 ms range. As latency rose into several hundred milliseconds and PRR degraded with density, the controller increasingly shifted toward uncertainty inflation, reduced trusted-neighbor sets, guarded fallback, and backstop readiness. At 50 vehicles/km², the exchange remained usable across nominal-to-moderate impairment; around 100-150 vehicles/km², performance became scenario-dependent; and the 250 vehicles/km² communication-envelope point should be interpreted as a diagnostic breakdown stress endpoint rather than a viable operating regime.

This boundary clarifies the role of the three baselines. Baseline A represents the strategic coordination layer most concretely defined in current deployment-oriented UTM architectures. The A-B2 comparison asks whether intent-first V2V can extend usable operation beyond strategic coordination alone under disturbances that would otherwise be absorbed through holding, replanning, or broad reauthorization. Baseline B1 represents tactical capability without deployable V2V constraints. The B1-B2 comparison isolates both the operational role of V2V-enabled neighborhood exchange and the cost of making that exchange deployable under authentication, bounded freshness, congestion, coverage limits, and degraded-mode behavior. Lower B2 raw throughput in some regimes should therefore be interpreted as part of the operational cost of deployable tactical coordination: the controller rejects invalid messages, discounts stale intent, and avoids overconfident coordination under partial information.

Future work should focus on the transition region between cooperative tactical operation and guarded fallback, especially congestion control, shared-resource admission and sequencing, integrity-aware prediction, cooperative perception under dense occlusion, and tighter integration between tactical planning and the onboard backstop.

*D. Platform, SWaP, and Spectrum Implications*

The evaluated implementation also shows that tactical V2V is a stack-level integration problem. The hardware tests demonstrate practical feasibility for the evaluated class of mid-size UAS: airborne nodes carried the V2V module, antenna, and GNSS while operating in direct air-to-air mode with no ground relay. This does not mean SWaP is universally solved across all vehicle classes. Rather, the relevant deployment burden includes radio and antenna integration, onboard scheduling, message encoding, authentication verification, time synchronization, logging, controller interfacing, and degraded-mode triggering. Operational systems will likely require tighter avionics or autonomy-stack integration than prototype experimental modules.

The spectrum implication is similarly practical. The evaluated stack used a 5.9 GHz ITS-band carrier as an experimental carrier for the studied tactical-neighborhood profile, not as a claim that this band is a sufficient aviation spectrum solution. Any candidate spectrum or channelization approach for tactical aerial V2V should therefore be evaluated against the full tactical exchange burden, not against awareness-only broadcast or sparse event-notification assumptions: periodic intent exchange, bursty event-triggered coordination, contingency signaling, cooperative-perception support, authentication metadata, congestion behavior, and degraded-mode signaling. A channel that supports occasional awareness broadcast may still be inadequate for freshness-bounded commitment, release, sequencing, or cooperative perception under density.

*E. Limitations*

Several limitations should be noted. First, the evaluated hardware, antenna configuration, and integration choices are representative of the tested stack, not universal. Other modules, power levels, antenna placements, vehicle geometries, and integration approaches may produce different PRR, latency, range, and congestion behavior.

Second, the city-anchor propagation model is a hardware-informed, geometry-aware stress model rather than a site-calibrated RF propagation study. Open-field hardware tests grounded LOS behavior, while city NLOS behavior was modeled using building geometry, material-aware attenuation, range scaling, and latency penalties rather than full electromagnetic ray tracing or direct urban RF calibration.

Third, the reported density breakpoints are scenario-derived and should not be interpreted as universal operational limits or regulatory thresholds. The evaluated stack includes densities up to 250 vehicles/km² to expose communication-envelope breakdown, but the practical deployable regime is substantially lower and depends on geometry, impairment, interaction structure, controller behavior, and vehicle dynamics.

Finally, dense shared-resource coordination remains only partially solved. The evaluated controller demonstrates the need for admission, sequencing, release, rejoin, and priority semantics, and it improves lower-density hotspot stability. However, the hardest dense terminal and burst-disruption cases still become degraded or unresolved. This paper therefore provides a controller-coupled characterization, protocol, and evaluation structure, not a complete certification basis, spectrum allocation, or interoperability standard.

## VIII. Conclusion

This paper presented an intent-first aerial V2V tactical neighborhood exchange for dense low-altitude airspace. The evaluated all-airborne C-V2X/PC5 stack combines periodic state and intent beacons, event-triggered coordination messages, authenticated freshness checks, and degraded-mode behavior to support the tactical layer between strategic deconfliction and last-resort collision avoidance.

The results show that V2V-enabled tactical coordination is most valuable in dense, disturbance-driven regimes where strategic reauthorization alone cannot absorb local change fast enough and collision avoidance should not become the routine traffic-management layer. Its value is mechanism-specific: it reduces stale belief, enables trusted coordination, suppresses false tactical inference, supports cooperative perception, structures shared-resource sequencing, and improves consistency among participating aircraft in mixed-equipage encounters. The evaluated stack remains bounded by density, impairment, observability, trust, and interaction complexity; it extends the tactical coordination envelope in lower-to-moderate stressed regimes but transitions toward guarded or backstop-heavy behavior as conditions worsen. The central contribution is a controller-coupled communication envelope for realizing the tactical layer on top of strategic coordination: it shows what intent-first V2V enables for local coordination and separation support, what operational costs deployable V2V imposes, and where conservative fallback becomes necessary.

## Acknowledgment

The author acknowledges Qualcomm Technologies, Inc. as the provider of prototype V2V communication hardware used in the experimental evaluation. The author also acknowledges computing resources provided by TACC Vista (NVIDIA GH100 Grace Hopper Superchip), Hugging Face Spaces and Compute Grants, and PSC Neocortex (Cerebras Wafer Scale Engine AI Accelerator).

## References

[1] ICAO, Unmanned Aircraft Systems Traffic Management (UTM) — A Common Framework With Core Principles for Global Harmonization, Edition 4, Montreal, QC, Canada: International Civil Aviation Organization, 2023.

[2] M. Johnson and J. Larrow, "UAS traffic management conflict management model," NASA, Tech. Rep., 2020. [Online]. Available: https://ntrs.nasa.gov/citations/20205002076

[3] Federal Aviation Administration, "Unmanned aircraft system (UAS) traffic management (UTM) concept of operations," v2.0, Washington, DC, USA, 2022.

[4] ASTM International, "Standard specification for UAS traffic management (UTM) UAS service supplier (USS) interoperability," ASTM F3548-21, West Conshohocken, PA, USA, 2021.

[5] M. R. Jamali et al., "UAS traffic management communications: The legacy of ADS-B, new establishment of Remote ID, or leverage of ADS-B-like systems?," Drones, vol. 6, no. 3, Art. no. 57, 2022, doi: 10.3390/drones6030057.

[6] Federal Aviation Administration, "Automatic dependent surveillance—broadcast (ADS-B) Out equipment performance requirements," 14 CFR §91.227, Federal Register, 2010.

[7] Federal Aviation Administration, "Remote identification of unmanned aircraft," 14 CFR Part 89, Federal Register, 2021.

[8] E. Vinogradov, F. Minucci, A. V. S. S. B. Kumar, S. Pollin, and E. Natalizio, "Reducing safe UAV separation distances with U2U communication and new Remote ID formats," 2022, arXiv:2209.13270.

[9] E. Vinogradov, A. V. S. S. B. Kumar, F. Minucci, S. Pollin, and E. Natalizio, "Remote ID for separation provision and multi-agent navigation," 2023, arXiv:2309.00843.

[10] H. Y. Ong and M. J. Kochenderfer, "Markov decision process-based distributed conflict resolution for drone air traffic management," J. Guid., Control, Dyn., vol. 40, no. 1, pp. 69–80, Jan. 2017, doi: 10.2514/1.G001822.

[11] S. Hoogendoorn, V. L. Knoop, H. Mahmassani, and S. Hoogendoorn-Lanser, "Game-theoretical approach to decentralized multi-drone conflict resolution and emergent traffic flow operations," 2023, arXiv:2308.01069.

[12] B. Vásárhelyi et al., "Decentralized traffic management of autonomous drones," Swarm Intell., vol. 18, 2024.

[13] J. Tordesillas and J. P. How, "MADER: Trajectory planner in multiagent and dynamic environments," IEEE Trans. Robot., vol. 38, no. 1, pp. 463–476, Feb. 2022, doi: 10.1109/TRO.2021.3080235.

[14] K. Kondo et al., "Robust MADER: Decentralized multiagent trajectory planner robust to communication delay in dynamic environments," IEEE Robot. Autom. Lett., vol. 8, no. 3, pp. 1687–1694, Mar. 2023, doi: 10.1109/LRA.2023.3236923.

[15] A. M. Khasawneh et al., "Latency analysis of drone-assisted C-V2X communications for basic safety and cooperative perception messages," Drones, vol. 8, no. 10, Art. no. 600, 2024, doi: 10.3390/drones8100600.

[16] S. Janji, "Deploying an aerial reconfigurable intelligent surface for vehicle-to-vehicle communications," in Proc. Krajowe Sympozjum Telekomun. Teleinf. (KRiT), 2024.

[17] D. Kavas-Torris, S. Y. Gelbal, M. R. Cantas, B. Aksun-Guvenc, and L. Guvenc, "V2X communication between connected and automated vehicles (CAVs) and unmanned aerial vehicles (UAVs)," Sensors, vol. 22, no. 22, Art. no. 8941, 2022, doi: 10.3390/s22228941.

[18] O. Berton et al., "Making detect and avoid a reality for UAS: Investigating leveraging 3GPP C-V2X technology for aerial applications," MITRE Engenuity, McLean, VA, USA, Doc. OG0074, 2023.

[19] G. Gür et al., "V2V UAS communications and use cases for advanced air mobility," in Proc. IEEE Conf. Standards Commun. Netw. (CSCN), 2024, pp. 103–107.

[20] K. Namuduri, J. S. Mandapaka, and M. F. Kidane, "The philosophy of UAS-to-UAS communications," in Proc. MOBILITY 2024: 14th Int. Conf. Mobile Services, Resour. Users, 2024, pp. 1–6.

[21] Z. McCorkendale, L. McCorkendale, and K. Namuduri, "Digital traffic lights: UAS collision avoidance strategy for advanced air mobility," Drones, vol. 8, no. 10, Art. no. 590, 2024, doi: 10.3390/drones8100590.

[22] J. S. Mandapaka, B. Dalloul, S. Hawkins, K. Namuduri, S. Nicoll, and K. Gambold, "Collision avoidance strategies for cooperative unmanned aircraft systems using vehicle-to-vehicle communications," in Proc. IEEE 97th Veh. Technol. Conf. (VTC2023-Spring), Florence, Italy, Jun. 2023, pp. 1–7, doi: 10.1109/VTC2023-Spring57618.2023.10199913.

[23] J. S. Mandapaka, L. McCorkendale, Z. McCorkendale, and K. Namuduri, "Collision avoidance strategies for advanced air mobility using UAS-to-UAS communications," Drone Syst. Appl., vol. 13, 2025, doi: 10.1139/dsa-2024-0044.

[24] RTCA, "Vehicle to vehicle communications white paper," RTCA, Washington, DC, USA, White Paper, Dec. 2022. [Online]. Available: https://www.rtca.org/wp-content/uploads/2023/01/V2V-White-Paper-Final.pdf

[25] IEEE Standards Association, "IEEE P1920.2: Standard for vehicle-to-vehicle communications for unmanned aircraft systems," Active Standards Project, Piscataway, NJ, USA.

[26] A. Aweiss et al., "Flight demonstration of unmanned aircraft system (UAS) traffic management (UTM) at technical capability level 3," in Proc. IEEE/AIAA 38th Digit. Avion. Syst. Conf. (DASC), San Diego, CA, USA, Sep. 2019, pp. 1–7, doi: 10.1109/DASC43569.2019.9081718.

[27] A. Chakrabarty, C. A. Ippolito, J. Baculi, K. S. Krishnakumar, and S. Hening, "Vehicle to vehicle (V2V) communication for collision avoidance for multi-copters flying in UTM-TCL4," in Proc. AIAA Scitech Forum, San Diego, CA, USA, Jan. 2019, doi: 10.2514/6.2019-0690.

[28] T. Yu, K. Araki, and K. Sakaguchi, "Full-duplex aerial communication system for multiple UAVs with directional antennas," 2022, arXiv:2202.00176.

[29] D. Mishra et al., "Cooperative cellular UAV-to-everything (C-U2X) communication based on 5G sidelink for UAV swarms," Comput. Commun., vol. 192, pp. 173–184, 2022.

[30] M. Varonen, "Using NR sidelink for UAV swarm internal communication," M.S. thesis, School Elect. Eng., Aalto Univ., Espoo, Finland, 2024.

[31] A. Gürses, J. Kesler, and M. L. Sichitiu, "Air-to-air channel characterization for UAV communications at 3.4 GHz," 2026, arXiv:2604.01582.

[32] J. Beuster et al., "Enhancing situational awareness in ISAC networks via drone swarms: A real-world channel sounding data set," in Proc. 28th Int. Workshop Smart Antennas (WSA), 2025, pp. 170–173.

[33] T.-W. Hsu et al., "Vehicle-to-vehicle based autonomous flight coordination control system for safer operation of unmanned aerial

vehicles," Drones, vol. 7, no. 11, Art. no. 669, 2023, doi: 10.3390/drones7110669.

[34] 3GPP, "Support of unmanned aerial systems (UAS) connectivity, identification and tracking; Stage 2," 3GPP TS 23.256, Release 18, Sophia Antipolis, France, 2024.

[35] R. Xu et al., "V2X-ViT: Vehicle-to-everything cooperative perception with vision transformer," in Proc. Eur. Conf. Comput. Vis. (ECCV), Tel Aviv, Israel, 2022, pp. 107–124.

[36] R. Xu et al., "OPV2V: An open benchmark dataset and fusion pipeline for perception with vehicle-to-vehicle communication," in Proc. IEEE Int. Conf. Robot. Autom. (ICRA), Philadelphia, PA, USA, 2022, pp. 2583–2589, doi: 10.1109/ICRA46639.2022.9812038.

[37] W. Zimmer et al., "TUMTraf V2X cooperative perception dataset," in Proc. IEEE/CVF Conf. Comput. Vis. Pattern Recognit. (CVPR), Seattle, WA, USA, 2024.

[38] J. Zhao et al., "CoMamba: Real-time cooperative perception unlocked with state space models," 2024, arXiv:2409.10699.

[39] Y. Wang et al., "CoFormerNet: A transformer-based fusion approach for enhanced vehicle-infrastructure cooperative perception," Sensors, vol. 24, no. 13, Art. no. 4101, 2024, doi: 10.3390/s24134101.

[40] L. Cui et al., "U2UData: A large-scale cooperative perception dataset for swarm UAVs autonomous flight," in Proc. ACM Int. Conf. Multimedia (MM), Melbourne, Australia, 2024.

[41] Q. Wei et al., "MCOP: Multi-UAV collaborative occupancy prediction," 2025, arXiv:2510.12679.

[42] L. Hou et al., "AGC-Drive: A large-scale dataset for real-world aerial-ground collaboration in driving scenarios," 2025, arXiv:2506.16371.

[43] X. Gao et al., "AirV2X: Unified air-ground vehicle-to-everything collaboration," 2025, arXiv:2506.19283.

[44] ETSI, "Intelligent transport systems (ITS); Vehicular communications; Basic set of applications; Cooperative awareness service," ETSI EN 302 637-2 V1.4.1, Sophia Antipolis, France, 2019.

[45] ETSI, "Intelligent transport systems (ITS); Vehicular communications; Basic set of applications; Decentralized environmental notification service," ETSI EN 302 637-3 V1.3.1, Sophia Antipolis, France, 2019.

[46] SAE International, "V2X communications message set dictionary," SAE J2735_202309, Warrendale, PA, USA, 2023.

[47] IEEE, "IEEE standard for wireless access in vehicular environments — Security services for applications and management messages," IEEE Std 1609.2-2022, Piscataway, NJ, USA, 2022.

[48] IEEE, "Certificate management interfaces for end entities," IEEE Std 1609.2.1-2022, Piscataway, NJ, USA, 2022.

[49] Y. Li, W. Chen, S. Peeta, and Y. Wang, "Platoon control of connected multi-vehicle systems under V2X communications: Design and experiments," IEEE Trans. Intell. Transp. Syst., vol. 21, no. 5, pp. 1891–1902, May 2020, doi: 10.1109/TITS.2019.2905039.

[50] S.-W. Kim et al., "Multivehicle cooperative driving using cooperative perception: Design and experimental validation," IEEE Trans. Intell. Transp. Syst., vol. 16, no. 2, pp. 663–680, Apr. 2015, doi: 10.1109/TITS.2014.2352853.

[51] X. Chen, S. Leng, J. He, L. Zhou, and H. Liu, "The upper bounds of cellular vehicle-to-vehicle communication latency for platoon-based autonomous driving," IEEE Trans. Intell. Transp. Syst., vol. 24, no. 7, pp. 6874–6887, Jul. 2023, doi: 10.1109/TITS.2023.3284574.

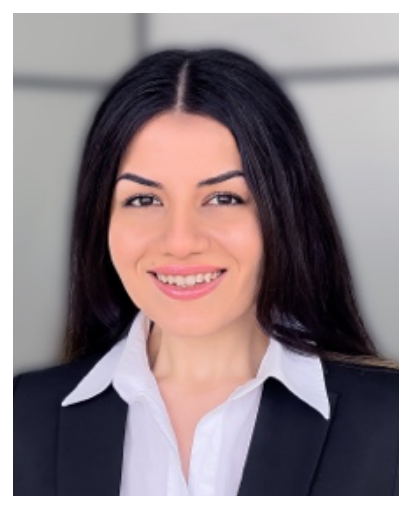

**Mehrnaz Sabet** is a Ph.D. candidate in the College of Computing and Information Science at Cornell University, Ithaca, NY, USA. She leads research on collaborative aerial autonomy in the Collaborative Technologies Lab, where she develops multi-agent machine learning and simulation systems for unmanned aircraft systems and human-autonomy teaming. She is the Principal Investigator of a NASA-funded multi-year project on AI-enabled traffic management for Advanced Air Mobility. Her research interests include aerial robotics, multi-agent autonomy, human-autonomy interaction, and autonomy safety validation. She has received research awards from NSF, NASA, and Cornell Engineering and serves in national leadership roles through the AIAA Human–Machine Teaming Technical Committee. She received the M.S. degree in information science from Cornell University, Ithaca, NY, USA.